\documentclass[10pt,journal,compsoc]{IEEEtran}
\ifCLASSOPTIONcompsoc
  \usepackage[nocompress]{cite}
\else
  \usepackage{cite}
\fi
\ifCLASSINFOpdf
\else
\fi
\usepackage{amsmath}
\usepackage{times}
\usepackage{epsfig}
\usepackage{graphicx}
\usepackage{amsmath}
\usepackage{amssymb}

\usepackage{tabularx}
\usepackage{booktabs}
\usepackage{multirow}

\newcommand{\etal}{\emph{et al.}}
\newcommand{\ie}{\emph{i.e.\ }}
\newcommand{\eg}{\emph{e.g.\ }}

\newcommand{\cf}{\emph{cf.\ }}

\newcommand{\argmax}{\operatornamewithlimits{arg\,max}}

\begin{document}
%
\title{A Hybrid RNN-HMM Approach for Weakly Supervised Temporal Action Segmentation}

%

%
%

\author{Hilde~Kuehne*,
        Alexander~Richard*,
        and~Juergen~Gall,~\IEEEmembership{Member,~IEEE}
\thanks{All authors are with the Institute
of Computer Science III, University of Bonn, Germany, *denotes equal 
contribution. E-mail: kuehne@iai.uni-bonn.de .}
}

\markboth{}%
{Kuehne \MakeLowercase{\textit{et al.}}: A Hybrid RNN-HMM Approach for Weakly Supervised Temporal Action Segmentation}
%


\IEEEtitleabstractindextext{%
\begin{abstract}

Action recognition has become a rapidly developing research field within the 
last decade.
But with the increasing demand for large scale data, the need of hand annotated data for the training becomes more and more impractical. One way to avoid frame-based human
annotation is the use of action order information to learn the respective action classes.  
In this context, we propose a hierarchical approach to address the problem of 
weakly supervised learning of human actions from ordered action labels by 
structuring recognition in a coarse-to-fine manner. Given a set of videos and an ordered list of the occurring actions, the task is to infer start and end frames of the related action classes within the video and 
to train the respective action classifiers without any need for hand labeled 
frame boundaries. We address this problem by combining a framewise RNN model 
with a coarse probabilistic inference. This combination allows for the temporal 
alignment of long sequences and thus, for an iterative training of both 
elements. While this system alone already generates good results, we show that 
the performance can be further improved by approximating the number of 
subactions to the characteristics of the different action classes as well as by 
the introduction of a regularizing length prior. 
The proposed system is evaluated on two benchmark datasets, the Breakfast and 
the Hollywood extended dataset, showing a competitive performance on various 
weak learning tasks such as temporal action segmentation and action alignment.

\end{abstract}

}

\maketitle

\IEEEdisplaynontitleabstractindextext

%
\IEEEpeerreviewmaketitle


\IEEEraisesectionheading{\section{Introduction}\label{sec:introduction}}
%
%
%
%
\IEEEPARstart{A}{ction} recognition has been a vivid and productive field within the last decade. So far research in this area is mainly focused on fully supervised classification of short, pre-clipped video snippets as the progress on the major benchmarks in this field shows~\cite{wang2013action, peng2014action, jain2015what, Wang2016temporal, carreira2017quo}.
But the collection of training data for such systems is usually a cost and time consuming process. Human annotators need to identify the specific classes in large video collections and manually mark the start and end frame of each corresponding action. This obviously makes it impractical to cover a larger amount of action classes and does not necessarily meet the preconditions of real live systems. This problem is further aggravated by the need for large scale training data for most deep learning approaches as shown by~\cite{carreira2017quo}. Additionally the costs of training data collection make it hard to acquire enough data to advance concepts beyond short clips \eg towards long term temporal models.

One way to address this issue is to give up on the need of frame-based annotation and to use only action labels and their ordering information to learn the respective action classes. This information is much easier to generate for human annotators, or can even be automatically derived from scripts~\cite{laptev08learning,marszalek09actions} or subtitles~\cite{alayrac16unsupervised}.
First attempts to address this kind of problem have been made by~\cite{bojanowski14weakly, alayrac16unsupervised, huang2016connectionist, kuehne2016weakly}.

In this context, we propose a hierarchical approach to address the problem of weakly supervised learning of human actions from transcripts. The method combines recognition in a coarse-to-fine manner. 
On the fine grained level, we use a discriminative representation of subactions, modeled by a recurrent neural network as \eg used by \cite{donahue15longterm, ng15beyond, singh16multistream, wu15modeling}. In our case, the RNN is used as basic recognition model as it provides robust classification of small temporal chunks. This allows to capture local temporal information. The RNN is supplemented by a coarse probabilistic model to allow for temporal alignment and inference over long sequences.

To bypass the difficulty of modeling long and complex action classes, we divide all actions into smaller building blocks.
Those subactions are eventually modeled within the RNN and later combined by the inference process. The usage of subactions allows
to distribute heterogeneous information of one action class over many subclasses and to capture characteristics such as the length of the overall action class. We show that automatically learning the number of subactions for each action class leads to a notable improved performance.

The obvious advantage of this kind of model is that it allows recognition of fine grained movements by still capturing mid and long temporal relations between frame responses. But the model is also especially suitable for the task of weak learning because it enforces a modular structure, as frame based responses are first combined to action classes and then to activity sequences. This allows for an iterative refinement of fine-grained and coarse recognition as well as an alternating adaptation of both elements. 

Our model is trained with an iterative procedure. Given the weakly supervised training data, an initial segmentation is generated
by uniformly distributing all actions among the video. For each action segment, all subactions are uniformly
distributed among the part of the video belonging to the corresponding action. This way, an initial alignment between video frames
and subactions is defined. In an iterative phase, the RNN is then trained on this alignment and used in combination with the
coarse model to infer new action segment boundaries. From those boundaries, we recompute the number of subactions needed for each action class, distribute them again among the frames aligned to the respective action, and repeat the training process until convergence.

To further improve the performance in this context, we extend the standard HMM formulation by the introduction of a state length prior during inference. The length prior serves as an additional regulation to balance the temporal dynamic of the system. The intuition underlying this concept is that actions are usually not only characterized by their specific movements, but also by the duration that is necessary to execute a certain task. One way to include this characteristic in the proposed system is to model the length of an action by the number of HMM states used to represent the action. But we found that depending on the observation prior, a number of states will aggregate all frames of an action during inference, thus undermining the original idea of representing variable length actions by adapting the number of states only. Therefore, the proposed length prior serves as an additional regularization factor to enforce a meaningful length of the single states. We will show that the length prior helps to prevent degenerated states during inference and thus to improve recognition accuracy in general.

We evaluate our approach on two common benchmark datasets, the Breakfast dataset~\cite{kuehne14language} and the Hollywood extended dataset \cite{bojanowski14weakly}, regarding two different tasks. The first task is temporal action segmentation, which refers to a combined segmentation and classification, where the test video is given without any further annotation. The second task is aligning a test video to a given order of actions, as proposed by Bojanowski \etal~\cite{bojanowski14weakly}. 

We regard those tasks with respect to different learning settings. First, in the weakly supervised scenario, we use only action labels and their ordering information to learn the respective action classes. Second, we extend the task to a semi supervised scenario by adding sparse frame-level annotation to the training data as shown in \cite{huang2016connectionist}. In this case, which we refer to as semi-supervised learning, a small fraction of the frames in the training set is annotated with the respective action class. It shows that, \eg on Breakfast,  using annotation of $0.25\%$ of all frames (on average one annotated frame per action instance) is enough to improve the overall accuracy significantly. To put those findings into context, we further evaluate the proposed system for the case of fully supervised recognition of temporal sequences. It shows that the sparse supervision on frame level is able to reach results comparable to full supervision. Additionally, for all evaluated cases the system is able to outperform any other state-of-the-art approach so far.

A preliminary version of this work has been published in \cite{richard2017weakly}. This work extends the previous approach by two main contributions. First, we extend the proposed model by the motivation and introduction of a length model. We show how the length model helps to  regularize the training and inference of the temporal model and that it outperforms the previous method as well as any other state-of-the-art approaches. We further modify the proposed training procedure to include sparse frame-level supervision to the system. This allows us to address the case of semi-supervised learning under sparse temporal supervision. Based on that, we not only show that the system is able to outperform any other system in this task, but we also demonstrate that the proposed length model helps not only for the case of weakly supervised learning, but also in case of semi- and fully supervised action recognition in general.



\section{Related work}
\label{sec:relatedWork}

Action recognition has come a long way within the last years, moving forward from highly tuned hand-crafted features as proposed by Wang \etal~\cite{wang2013dense, wang2013action} towards learned features and temporal connections as \eg proposed by Simonyan and Zisserman \cite{simonyan2014twostream}, Feichtenhofer \etal~\cite{Feichtenhofer2016Convolutional} and Wang \etal~\cite{Wang2016temporal}. Recent approaches show that results for classical action recognition are nearly ceiling on standard datasets \cite{carreira2017quo, girdhar2017actionVLAD}. Alternative approaches focus on the learning of temporal sequences without full supervision. In the following, we will first give an overview of concepts for weakly supervised learning from sequential data as well as weakly supervised fine-tuning with pretrained models. Finally, we will review different scenarios in the context of duration modeling for temporal sequences.

\subsection{Weakly supervised learning from structured sequences}
Compared to classical action recognition, the problem of weakly supervised learning of actions is a rather new topic. First works in this field, proposed by Laptev \etal~\cite{laptev08learning} and Marszalek \etal~\cite{marszalek09actions}, focus on mining training samples from movie scripts. They extract class samples based on the respective text passages and use those snippets for training without applying a dedicated temporal alignment of the action within the extracted clips. 
Attempts for learning action classes including temporal alignment on weakly annotated data are made by Duchenne \etal~\cite{duchenne09automatic}. Here, it is assumed that video clips contain only one class and the task is to temporally segment frames containing the relevant action from the background activities. The temporal alignment is thus interpreted as a binary clustering problem, separating temporal snippets containing the action class from the background segments. The clustering problem is formulated as a minimization of a discriminative cost function. This problem formulation is extended by Bojanowski \etal~\cite{bojanowski14weakly} also introducing the Hollywood extended dataset. The weak learning is formulated as a temporal assignment problem. Given a set of videos and the action order of each video, the task is to assign the respective class to each frame, thus to infer the respective action boundaries. The authors propose a discriminative clustering model using temporal ordering constraints to combine classification of each action and their temporal localization in each video clip. They propose the usage of the Frank-Wolfe algorithm to solve the convex minimization problem. This method has been adopted by Alayrac \etal~\cite{alayrac16unsupervised} for unsupervised learning of task and story lines from instructional video.
Another approach for weakly supervised learning from temporally ordered action lists is introduced by Huang \etal~\cite{huang2016connectionist}. Inspired by CTC models in speech recognition \cite{Graves2006CTC}, they use extended connectionist temporal classification and introduce a visual similarity measures to prevent the CTC framework from degeneration and to enforce visually consistent paths. 
A different way, also lending on the concept of speech recognition, is proposed by Kuehne \etal~\cite{kuehne2016weakly}. Here, actions are modeled by hidden Markov models (HMMs) with the aim to maximize the probability of training sequences being generated by the HMMs, by iteratively inferring the segmentation boundaries for each video and using the new segmentation to re-estimate the model. The last two approaches were both evaluated on the Hollywood extended as well as on the Breakfast dataset, thus, these two datasets are also used for the evaluation of the proposed framework. 
Another idea is proposed by Ding and Xu \cite{Ding2018weakly} using a temporal
convolutional feature pyramid network (TCFPN), an adaption of the encoder-decoder temporal convolutional neural networks (TCN) \cite{lea2017temporal}, for frame-wise classification in combination with an iterative soft boundary assignment for the action sequence alignment. During training, the alignment of the sequences to the transcripts is refined by an insertion strategy, which means that one class instance is represented by multiple successive instances of the same class, allowing the temporal receptive field of the TCN to extend in the temporal domain.  


\subsection{Weakly supervised fine-tuning}
Other works focus on fine-tuning pretrained models by detecting actions in untrimmed videos. Here, usually pretrained networks are used to detect unseen actions in an untrimmed training set and networks are fine-tuned with respect to the detected instances. Note that classes in the datasets used for pretraining, such as UCF101\cite{soomro2012UCF101}, Sports1M\cite{karpathy2014large} or Kinetics\cite{kay2017kinetics}, can overlap with the classes to search for in the untrimmed videos \eg in case of the Thumos action detection task \cite{Jiang2014THUMOS}. There is further the information given, which classes appear in the untrimmed videos, but not when or how often they appear.  
This task was first addresses by Wang \etal~\cite{wang2017untrimmed} by selecting clip proposals from a set of untrimmed training videos to learn actions without exact boundary annotation. Nguyen \etal~\cite{Nguyen2018weakly} propose a combination of attention weights and temporal class activation maps for the task. 

\subsection{Weakly supervised approaches in other domains}
Koller \etal~\cite{koller16deephand} integrate CNNs with hidden Markov models to learn sign language hand shapes based on a single frame CNN model from weakly annotated data. They extend the proposed single frame model by including LSTMs for temporal correlation in \cite{koller2017resign}.
A more speech related task is also proposed by Malmaud \etal~\cite{malmaud15what}, trying to align recipe steps to automatically generated speech transcripts from cooking videos. They use a hybrid HMM model in combination with a CNN based visual food detector to align a sequence of instructions, \eg from textual recipes, to a video of someone carrying out a task.
Gan \etal~\cite{gan2016webly} learn action classes from web images and videos retrieved by specific search queries. They match images and video frames and use a regularization over the selected video frames to balance the matching procedure. 
\cite{sun15temporal} also takes weak video labels and noisy image labels as input and generates localized action frames as output. The localized action frames are used to train action recognition models. 
Yan \etal~\cite{Yan2017weakly} use video-level tags for weakly-supervised actor-action segmentation using a multi-task ranking model to select representative supervoxels for actors and their respective actions.
Finally, \cite{wu15watch} propose an unsupervised technique to derive action classes from RGB-D videos using Gibbs sampling for learning long activities from basic action words.

\subsection{Length modeling for temporal sequences}
Bojanovski \etal~\cite{Bojanowski2015weakly} exploit a length model for weakly supervised video-to-text alignment. They regularize the length of the video segments that are supposed to be aligned with the respective description. Probably closest to the here proposed length prior is the work of Richard \etal~\cite{richard2016temporal}, which introduces a length model that depends on the overall length of an action with respect to the mean length of the recognized class. This model is able to capture any discrete probability distribution and penalizes too short as well as too long sequences. The problem of this formulation is that the first-order dependence of the model leads to a quadratic runtime in the inference and therefore becomes unfeasible for longer temporal video data. 
The modeling of temporal duration has also a long tradition in the context of speech processing and has been used \eg in \cite{VASEGHI199531}. Since then it has been used in different contexts such as general modeling in case of explicit state duration HMMs~\cite{Dewar2012Inference} but also for speech synthesis \cite{Zen2007Hidden} or the generation and decoding of temporal sequences, such as music \cite{Narimatsu2017State}.




\section{Task description}
\label{sec:taskDescription}

\subsection{Learning from transcripts only}

In contrast to fully supervised action detection or segmentation approaches, where frame based annotation is available, weakly supervised learning is based on an ordered list of the actions occurring in the video. 
A video of the activity ``Making tea'' might consist of taking a cup, putting the teabag in it, and pouring water into the cup.
In a fully supervised task, a temporal annotation of each action start and end time would be available for training, \eg in form of 

\begin{center}
    \texttt{\phantom{00}0 - 21: take\_cup\phantom{000}} \\
    \texttt{22 - 68: add\_teabag} \\
    \texttt{69 - 73: pour\_water}. \\
\end{center}

In our weakly supervised setup, all videos are just labeled with their ordered action sequence given as 

\begin{center}
\texttt{take\_cup, add\_teabag, pour\_water}.
\end{center}

As this information is available for each video and as long as all actions appear at least once in different contexts, it is possible to infer the related action boundaries without frame-based ground truth information, in our case by choosing the related action representation in a way that they maximize the probability that the sequences were generated by the respective models.

Note that this also formulates the necessary preconditions of the overall system, namely the fact that it needs the order of all actions as they appear in the sequences of the training set and that all actions need to appear at least with two different predecessors and successors. This constrain is necessary as we want to maximize the probability that the sequences of the training data are generated by models trained on a set of boundary assumptions. If \eg two actions always occur together and in the same order, the combined probability for both models will be the same, no matter where a boundary point is set between those two actions. Only if one action appears in a different context, \ie with different predecessors and successors, we are able to maximize the overall probability of the system with respect to different boundaries.

\subsection{Learning from transcripts including sparse frame-level annotation}

A slightly modified version of this problem is the learning from transcripts including sparse frame-level annotation.  Here, again, the transcripts are provided as described, but additionally a fraction of the frames is also annotated with their respective ground truth label. This setting is motivated by the idea that single frame labels are usually easier to acquire than a full action segmentation. It requires only to look at a few frames without the need to watch the whole video. Such annotations can be collected \eg  via captcha tasks, Mechanical Turk or in similar settings. In this case, the annotation information for the video ``Making tea'' might look as follows 

\begin{center}
\item \texttt{take\_cup, add\_teabag, pour\_water} \\
\texttt{frame 3: take\_cup} \\
\texttt{frame 65: add\_teabag}. 
\end{center}

The frame information does not refer to any action boundaries. It only indicates at which position in the video a certain action occurs. Thus, the task is still to infer the related action boundaries, but under the additional constraint of matching the annotated frames.


\section{System overview}
\label{sec:systemOverview}

As sequential actions are naturally composed of hierarchical movements and actions at different levels of temporal granularity, we follow the idea of a hierarchical action model and adapt it for the case of weak learning of human actions in video data.

At top level, we model each temporal sequence as a combination of basic actions. This can be an activity, as e.g. ``Making tea'' which would be made up of the actions ``take cup'', ``add teabag'' and ``pour water''. Each of those actions is represented by a respective probabilistic graph model, in this case an HMM, which models each action as a combination of subactions. Intuitively, the idea of subactions is based on the fact that, e.g. an action such as ``take cup'' consists of multiple movements like ``move hand towards cup'', ``grab cup'', ``move cup towards body'' and ``release cup''. 

The proposed model captures those implicitly available but not explicitly annotated subactions as latent variables by the states in the graph. In order to build the state graph, it is not necessary to know the true number or label of the possible subactions. Instead, we set the number of subactions relative to the length of the corresponding action and update this factor as part of the training. Thus subactions at the beginning of an action  capture motion patterns typical for that phase, as e.g. for ``take cup'' the first subactions comprise elements such as ``move hand towards cup''. To ensure the sequential peculiarity of human actions within the state graph, we use a feed-forward topology, allowing only self-transition or transitions to the next state. We also show that this characteristic can be further supported by introducing a state specific length model to regularize the duration of each state during inference.

In the following, we describe the proposed framework in detail, starting with the formal definition of the hierarchical action model. After that, we discuss the different elements of our model, the fine-graind subaction classification and the length prior in detail. Next, we describe the inference and training procedure for the weak as well as for the semi supervised case and close with a discussion of the chosen stop criterion.

\subsection{Hierarchical action model}
\label{sec:actionModel}

As already stated, our training data consists of a set of videos and their respective transcripts, indicating the occurring actions in the correct order. 
Formally, we can assume the training data is a set of tupels $ (\mathbf{x}_1^T, \mathbf{a}_1^N ) $,
where $ \mathbf{x}_1^T = (x_1, \dots, x_T) $ are framewise features of a video with $ T $ frames and
$ \mathbf{a}_1^N $ is an ordered sequence $ (a_1, \dots, a_N) $ of actions occurring
in the video. The segmentation of the video is defined by the mapping
\begin{align}
    n(t): \{1,\dots,T\} \mapsto \{1,\dots,N\}
    \label{mapping}
\end{align}
that assigns an action segment index to each frame. 
Initially, this can simply be a linear segmentation of the
provided actions, see Figure~\ref{fig:train}a.
The likelihood of an action sequence $ \mathbf{x}_1^T $ given the action transcripts
$ \mathbf{a}_1^N $ is then defined as
\begin{align}
    p(\mathbf{x}_1^T | \mathbf{a}_1^N) := \prod_{t=1}^T p\big(x_t|a_{n(t)}\big),
\end{align}
where $ p(x_t|a_{n(t)}) $ are probabilities of frame $ x_t $ being generated by the action $ a_{n(t)} $.

The action classes usually describe longer, task-oriented procedures that naturally consist of more than one significant movement and we want to efficiently capture those characteristics. We model each action as a sequential combination of subactions.
For each action class $ a $, a set of subactions $ s_1^{(a)},\dots,s_{K_a}^{(a)} $ is defined. The number $ K_a $ is initially estimated by a heuristic and refined during the optimization process.
Practically, this means that we subdivide the original long action classes into a set of smaller subactions. As subactions are obviously not defined by the given ordered action sequences, we treat them as latent variables that need to be learned by the model. 
In the following system description, we assume that the subaction frame boundaries are known, \eg from previous iterations or from an initial uniform segmentation (see Figure~\ref{fig:train}b), and discuss the inference of concrete boundaries in Section \ref{sec:inference}.

In order to combine the fine grained subactions to action sequences, a hidden Markov model
$ \mathcal{H}_a $ for each action $ a $ is defined. The HMM ensures that subactions
only occur in the correct ordering, \ie that $ s_i^{(a)} \prec s_j^{(a)} $ for $ i \leq j $.
More precisely, let
\begin{align}
    s(t): \{1, \dots, T\} \mapsto \{s_1^{(a_1)},\dots,s_{K_{a_N}}^{(a_N)}\}
\end{align}
be the known mapping from video frames to the subactions of the ordered action sequence
$ \mathbf{a}_1^N $. This is basically the same mapping as the one in Equation~\eqref{mapping}
but on subaction level rather than on action level.
When going from one frame to the next, we only allow to assign either the same subaction or
the next subaction, so if at frame $ t $ the assigned subaction is $ s(t) = s^{(a)}_i $,
then at frame $ t + 1 $ either $ s(t+1) = s^{(a)}_i $ or $ s(t+1) = s^{(a)}_{i+1} $. In the following, we will denote $ s(t) $ by the subscript $ s_t $ for better readability.

The likelihood of an action sequence $ \mathbf{x}_1^T $ given the action transcripts
$ \mathbf{a}_1^N $ is then given by
\begin{align}
    p(\mathbf{x}_1^T | \mathbf{s}_1^T) :=
        \prod_{t=1}^T p\big(x_t|s_t\big) \cdot p\big(s_t|s_{t-1}\big),
        \label{coarseModel}
\end{align}
where $ p(x_t|s) $ are probabilities computed by the fine-grained subaction model, see Section~\ref{sec:finegrained}.
As defined by the feed forward model, the transition probabilities  $p(s_t|s_{t-1})$ can model a self-transition or a transition from subaction $ s' $ to subaction $ s $. In both cases, we compute the relative frequencies of how often the transition $ s' \rightarrow s $ occurs by regarding the $ s_t $-mappings of all training videos.

\subsection{Fine-grained Subaction Model}
\label{sec:finegrained}

\begin{figure}[tb]
    \centering
    \includegraphics{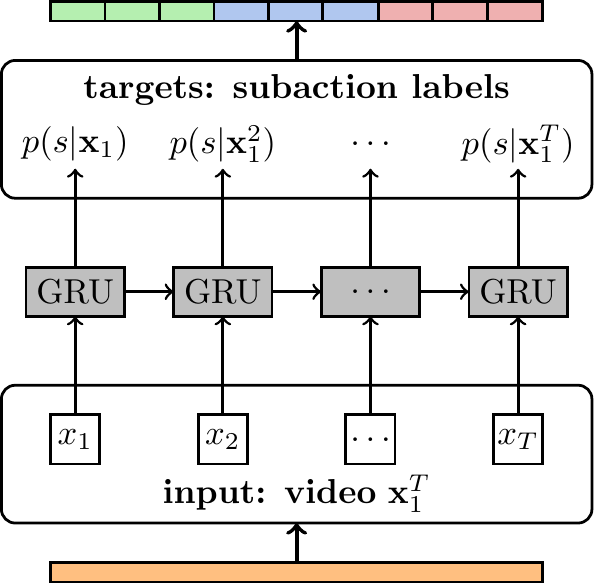}
    \caption{RNN using gated recurrent units with framewise video features as input.
             At each frame, the network outputs a probability for each possible subaction
             while considering the temporal context of the video by the preceding frames.}
    \label{fig:rnntrain}
\end{figure}

For the classification of fine-grained subactions, we use an RNN with a single hidden layer of gated recurrent units (GRUs) \cite{cho2014translation}, a simplified version of LSTMs that shows comparable performance~\cite{jozefowicz15empirical,chung2014empirical} also in case of video classification~\cite{ballas16delving}, but has less parameters than an LSTM unit. The network is shown in Figure \ref{fig:rnntrain}. 
For each frame, it predicts a probability distribution over all subactions, while the recurrent structure of the network allows to incorporate local temporal context. Since the RNN generates a posterior distribution
$ p(s|x_t) $ but our coarse model deals with subaction-conditional probabilities,
we use Bayes' rule to transform the network output to
\begin{align}
    p(x_t|s) = \mathrm{const} \cdot \frac{p(s|x_t)}{p(s)}, \label{bayes}
\end{align}
and thus allows for a direct usage of the distributions generated by the recurrent network in Equation~\eqref{coarseModel}.

As recurrent neural networks are usually trained using back propagation through time
(BPTT) \cite{werbos1990backpropagation}, which requires to process the whole sequence in a forward
and backward pass and a video can be very long and may easily exceed $ 10,000 $ frames,
the computation time per minibatch can be extremely high. We therefore adapt the training procedure by using small chunks around each video frame. They can be efficiently processed with a reasonably large minibatch size in order to
enable efficient RNN training on long videos. For each
frame $ t $, we create a chunk over $ \mathbf{x}[t - 20, t] $ and forward it through
the RNN. While this practically increases the amount of data that needs to be
processed by a factor of $ 20 $, only short sequences need to be forwarded at once and we
benefit from a high degree of parallelism and comparable large minibatch size.


\subsection{Inference}
\label{sec:inference}

To make use of the so far computed fine-grained subaction probabilities for long term recognition, we need to combine them over time to derive the underlying action classes as well as the overall temporal sequence. To do so, we use a temporal inference model, based on the hierarchical model formulation of our system. We will now discuss the inference on action and video level. Here different constraints can be applied. We first discuss the case of inference constrained by a grammar. Second, we consider the inference with given transcripts as a special case of a grammar with just one valid path. Third, we extend both modalities for sparse frame-level annotation.

\begin{figure}[tb]
    \centering
    \includegraphics[width=0.45\textwidth]{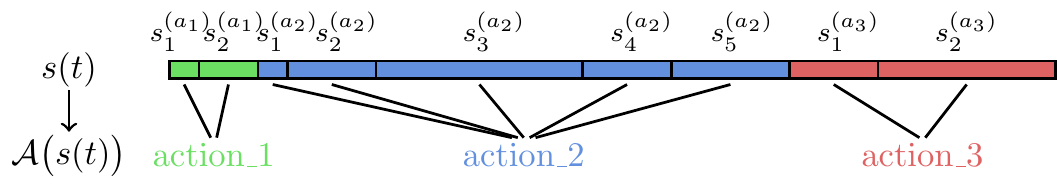}
    \caption{Example for the extractor function $ \mathcal{A} $. During inference, a frame-to-subaction alignment  $ s_t $ is found. To compute the respective unique action sequence. the extractor function maps the subactions back to its respective action classes.  }
    \label{fig:extractor}
\end{figure}

\textit{Inference with grammar.}
Given a video $ \mathbf{x}_1^T $ we want to find the most likely action sequence
\begin{align}
    \mathbf{\hat a}_1^N = \argmax_{\mathbf{a}_1^N} p(\mathbf{x}_1^T | \mathbf{a}_1^N)
    \label{optSequence}
\end{align}
as well as the corresponding frame alignment. 
In order to limit the amount of action sequences to optimize over, a context-free grammar $ \mathcal{G} $ is created by accumulating all action transcripts seen in the training data as in \cite{kuehne2016weakly}. 
Instead of finding the optimal action
sequence directly, the inference can equivalently be performed over all possible
frame-to-subaction alignments $ s_t $ that are consistent with $ \mathcal{G} $.
Consistent means that the unique action sequence defined by $ s_t $ is generated
by $ \mathcal{G} $. 

To receive the respective action sequence $ \mathbf{\hat a}_1^N $, we need to map the subaction output back to their original action classes. To this end, we define an extractor function $ \mathcal{A}: s(t) \mapsto \mathbf{a}_1^N $
that maps the frame-to-subaction alignment $ s_t $ to its action sequence, see Figure~\ref{fig:extractor}
for an illustration. Then, Equation~\eqref{optSequence} can be rewritten as
\begin{align}
    \mathbf{\hat a}_1^N = \argmax_{s_t: \mathcal{A}(s_t) \in \mathcal{L}(\mathcal{G})}
                          \Big\{ \prod_{t=1}^T p\big(x_t|s_t\big) \cdot p\big(s_t|s_{t-1}\big) \Big\},
                          \label{optAlignment}
\end{align}
where $ \mathcal{L}(\mathcal{G}) $ is the set of all possible action sequences that can be generated
by $ \mathcal{G} $. Equation~\eqref{optAlignment} can be solved efficiently using a Viterbi algorithm
if the grammar is context-free, see \eg \cite{jurafsky1995using}.

Note that we have two types of transitions, namely transitions between subactions of the same class, \ie \mbox{$\mathcal{A}(s_{t-1}) = \mathcal{A}(s_t)$}, as well as transitions between two action classes, \ie \mbox{$\mathcal{A}(s_{t-1}) \neq \mathcal{A}(s_t)$}. 
The transition probability $p(s_t|s_{t-1})$ for two subactions of the same action class can be computed from the current alignment of the training data. As we use a feed-forward model, we only allow self-transitions, \ie $s_t=s_{t-1}$, or transitions to the next state. The probability $p(s_t|s_{t-1})$ is computed by counting the respective number of transitions for the given training alignment. The transition probability between actions is $1$ if the transition encodes a viable path in the grammar and $0$ if not.

\textit{Inference with transcripts.}
\label{sec:alignmentWithTranscripts}
For training as well as for the task of aligning videos to a given ordered action sequence, the sequence of occurring actions $ \mathbf{a}_1^N $ is already known and only the best frame alignment to a single sequence needs to be inferred. By defining a grammar that generates the given action sequence $ \mathbf{a}_1^N $ only, this alignment task can be solved using Equation~\eqref{optAlignment}.

\textit{Inference with transcripts and sparse frame-level annotation.}
\label{sec:alignmentWithFrameLabels}
\begin{figure}[tb]
    \centering
    \includegraphics[width=0.45\textwidth]{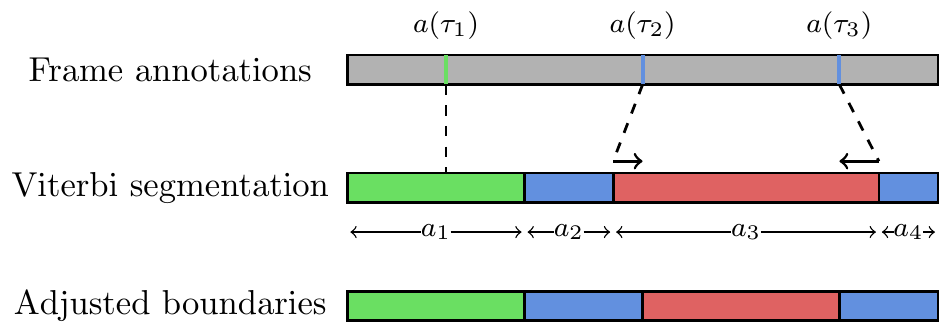}
    \caption{Boundary adjustment for semi-supervised training with sparse frame annotation. If the annotated frames are not consistent with the result after Viterbi decoding, the segmentation needs to be adjusted to fit the annotated frames. This also includes the association of the annotated frames to the respective segments. In this example, $ a(\tau_2) $ and $ a(\tau_3) $ both belong to the same class and could be associated to segment $ a_2 $ and $ a_4 $. Using a dynamic warping approach, boundary shifts are chosen to be as small as possible.
    }
    \label{fig:annotation}
\end{figure}
For the case that not only the transcripts but also sparse frame level annotation is available, the additional information is incorporated as additional guiding points, see Figure~\ref{fig:annotation} for an illustration. The problem  is that the guiding points only comprise the label for a specific frame, but do not include a matching of this frame to a certain segment in the transcripts. Thus, if the respective action occurs more than once, the frame can be assigned either way. To solve this assignment problem, we use a dynamic programming approach that assigns all labeled frames to their next matching action segment. We formulate a distance function of the inferred sequence $ \mathbf{a}_1^N $ and an ordered set of $ F $ frame labels $(\tau, a(\tau))$ which consists of frames indices $\tau$ and the respective class label  $a(\tau)$. We further denote the respective start and end frame of the $ n $-th segment of the Viterbi segmentation by $ start(a_n) $ and $ end(a_n) $. We assume that the respective segments have been found by the Viterbi algorithm as described before. The distance function between the current alignment and the sparse frame-level annotation is then given by

\begin{equation}
    d(\tau, a_n) = \begin{cases}
               0, & a(\tau) = a_n \ \land \\
                  & \tau \in [start(a_n) , end(a_n)], \\
               start(a_n) - \tau, & a(\tau) = a_n \ \land \\
                                    & \tau < start(a_n), \\
               \tau - end(a_n), & a(\tau) = a_n \ \land \\
                                  & \tau > end(a_n) \\
               \infty , & a(\tau) \neq a_n,\\
            \end{cases}
\end{equation}
\ie if the annotated frame $ \tau $ lies within the $ n $-th inferred segment and has the same label, $ a(\tau) = a_n $, the distance is zero (\eg $ a(\tau_1) $ and $ a_1 $ in Figure~\ref{fig:annotation}). If the annotated frame lies outside segment $ n $ but has the same action label, the distance is the number of frames it needs to be moved to lie within the $ n $-th segment (\eg $ a(\tau_2) $ and $ a_2 $ in Figure~\ref{fig:annotation}). If the annotated frame has another label than segment $ n $, \ie $ a(\tau) \neq a_n $, the distance is infinity (\eg $ a(\tau_2) $ and $ a_3 $ in Figure~\ref{fig:annotation}).

We minimize the respective distance function over all annotations for each video:

\begin{align}
    \min_{n(\tau): \{1,\dots,F\} \mapsto \{1,\dots,N\}} \left \{ \sum_{i=1}^{F} d( \tau_i, a_{n(\tau_i)}) \right \},
\end{align}
where $ F $ indicates the total number of framewise annotations.
The minimization is reached by a dynamic warping approach. The resulting mapping $ n(\tau) $ from labeled frames to action segments provides the assignment between frame annotations and Viterbi segments that requires the boundaries to be moved as little as possible.

For all setups with sparse frame-level supervision, we follow a two step procedure. Given the training videos as well as the respective transcripts and labels, we first infer the best path based on the transcripts and in a second step align the resulting boundaries with the given labeled frames.


\subsection{Length prior}
\label{sec:lengthprior}

\begin{figure}[tb]
    \centering
    \includegraphics{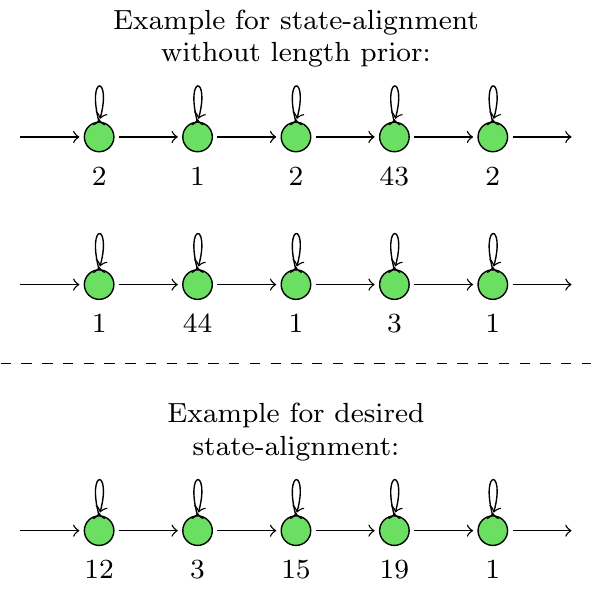}
    \caption{Example of state alignment for two instances of the same action as they are usually produced by the system without length prior and of an instance showing the intended state alignment. In the first two cases the HMM does not model the temporal progression, but rather uses the subaction states to distinguish between different action appearances. }
    \label{fig:HMM_state_example}
\end{figure}

We further add a state specific length prior as an additional regularization factor to our state model. The length prior serves in this case as a temporal decay model that rewards the model at the beginning of a new state to stay in this state for a certain amount of frames and punishes the model if it stays too long in the same state.
The idea is motivated by the observation that, without length prior, the system tends to skip many states, only remaining in those states for one or two frames. An example of this behaviour is shown in Figure \ref{fig:HMM_state_example}. It is clear to see that the HMMs in this case do not model any temporal progression. To evaluate this behaviour further, we counted the so called skip states, \ie states which are only assigned to one frame, for the case without and with length prior. We observe that without length prior $74.6\%$ of all processed states can be counted as skip states whereas the introduction of the length prior, \eg for the best performing configuration reported in Section \ref{sec:evaluation}, reduces the amount of skip states to $61.9\%$.

Formally, the length prior is a function of the duration of a state $ s $ at position $ t $ in the overall state-to-frame alignment, \ie a function of the length $l_t(s_t)$ that captures how long the model already remained in one state at time $ t $. The length function is defined recursively as

\begin{align}
    l_t(s_t) = \begin{cases}
               l_{t-1}(s_t)+1,& \text{if } s_t = s_{t-1}\\
               1, & \text{otherwise}\\
            \end{cases}.
\end{align}

The prior function $p\big(l_t(s_t)|s_t\big)$ models the decay factor based on the mean length $\mathrm{len}(s_t)$ of the respective state. The mean length is given by the average length of each state computed as
\begin{align}
    \mathrm{len}(s) = \frac{\text{number of frames aligned to } s}
                           {\text{number of }s\text{-instances}}.
                           \label{equ:meanLength}
\end{align}

An example for such a decay function can be a half Gaussian function defined as

\begin{align}
    \tilde p \big(l_t(s_t)|s_t\big) = e^{-\frac{(l_t(s_t)-\mu)^2}{\sigma^2}}
\end{align}
with $\mu = 0$ and $\sigma = \mathrm{len}(s_t)$, see Figure~\ref{fig:length_model} and the Appendix for more decay functions.

\subsection{Decoding with Length prior}
\label{sec:lengthprior_decoding}

The Viterbi decoding is a recursive function that computes the best path by maximizing the probability of all recent paths, encoded as a set of states, in combination with the probability of the current states.
More formally, we define the recursive function $Q$ to encode the maximum probability for a state path $s$ at time $t$. In the following, we define $s'$ to be a possible predecessor state at time $t-1$ and $ Q $ at $ (t-1,s') $ is the maximum probability of the HMM state path up to time $ t-1 $. The value of $ Q $ at $ (t,s) $ is then computed by selecting the HMM state path up to time $ t $ that ends in state $ s $ that maximizes the probability over all predecessor paths multiplied by the current observation probability $p(x_t|s)$ and the transition probability from the previous to the current state $p(s|s')$. 
This recursive equation solves the maximization from Equation~\eqref{optAlignment},
\begin{align}
   \tilde Q(t,s) = \max_{s'} \big\{ Q(t-1, s') \cdot p(x_t|s) \cdot p(s|s') \big\}.
    \label{equ:viterbi_rec}
\end{align}

Consequently, the overall best path by means of Equation~\eqref{optAlignment} is the best path ending at time $ T $, \ie the path with the score $ \max_s Q(T, s) $, see also~\cite{ney1999dynamic} for details.

We propose to use the length prior $ p \big( l_t(s_t) | s_t \big) $ as a regularizer during the Viterbi decoding to prevent hypotheses that stay in the same HMM state for too long. 
To this end we add the length prior to the overall decoding formulation
\begin{align}
    \mathbf{\hat a}_1^N = \argmax_{s_t: \mathcal{A}(s_t) \in \mathcal{L}(\mathcal{G})}
                          \Big \{ \prod_{t=1}^T p\big(x_t|s_t\big) \cdot p\big(s_t|s_{t-1}\big) \cdot p \big(l_t(s_t)|s_t\big) \Big \},
                          \label{optAlignment_wlength}
\end{align}

As the Viterbi decoding is carried out recursively, \cf Equation~\eqref{equ:viterbi_rec}, the previously multiplied length factor needs to be replaced by the current length factor when going from frame $ t-1 $ to frame $ t $. Therefore the overall prior is defined as
\begin{align}
    p \big( l_t(s_t) | s_t \big) = \frac{\tilde p \big( l_t(s_t) | s_t \big)}
                                          {\tilde p \big( l_{t-1}(s_t) | s_{t-1} \big)},
\end{align}
where we set $ \tilde p \big( l_{t-1}(s_t) | s_{t-1} \big) = 1 $ if $ s_t \neq s_{t-1} $.

In order to incorporate a regularization on the length, we modify the recursive equation by multiplying the length prior,
\begin{align}
    \tilde Q(t,s) = \max_{s'} \big\{ Q(t-1, s') & \cdot p(x_t|s) \cdot p(s|s') \nonumber \\
                                                & \cdot p \big(l_t(s_t)|s_t\big) \big\}.
    \label{viterbiRegularized}
\end{align}

Note that the length prior here works differently from context dependent length models as e.g. proposed in \cite{richard2016temporal}. Such length models assume a first-order dependence on the ending times of the single states. Including such a length model might allow for an even more precise length modeling but would require the Viterbi decoding to run not only over all time frames but also over all possible lengths, which would increase the runtime from linear in the frames, \ie $ \mathcal{O}(T) $, to quadratic in the frames, \ie $ \mathcal{O}(T^2) $. Since videos are usually long ($ T \gg 1000 $), this quickly becomes infeasible to compute. 

The main advantage of the here proposed length prior is that it only depends on the current length $l_t(s_t)$, which is also recursively defined and keeps the overall runtime linear.
This formulation requires the length prior to be a monotonically decreasing function. Consider a path that, at time $ t $, has just changed to state $ s $. A non-monotonous function such as a classical Poisson distribution would penalize such a path strongly, although the path may turn out very good if it stays in state $ s $ in the future. Using monotonous length priors avoids this problem and only paths that are in a certain state $ s $ for too long are penalized.
We define functions such as a \textit{half Poisson}, where the probability is constant up to the peak of the original Poisson distribution and then decreases as usual. Similarly, we propose a \textit{half Gaussian} that is centered at zero and decreases as the length increases like a Gaussian distribution, see Figure~\ref{fig:length_model} and the Appendix for details.

We evaluate the four different models, namely a box function, a linear decreasing function, a half Poisson, and a half Gaussian distribution in Section~\ref{sec:eval_lengthprior}.

\begin{figure*}[tb]
    \begin{center}
    \includegraphics[width=\textwidth]{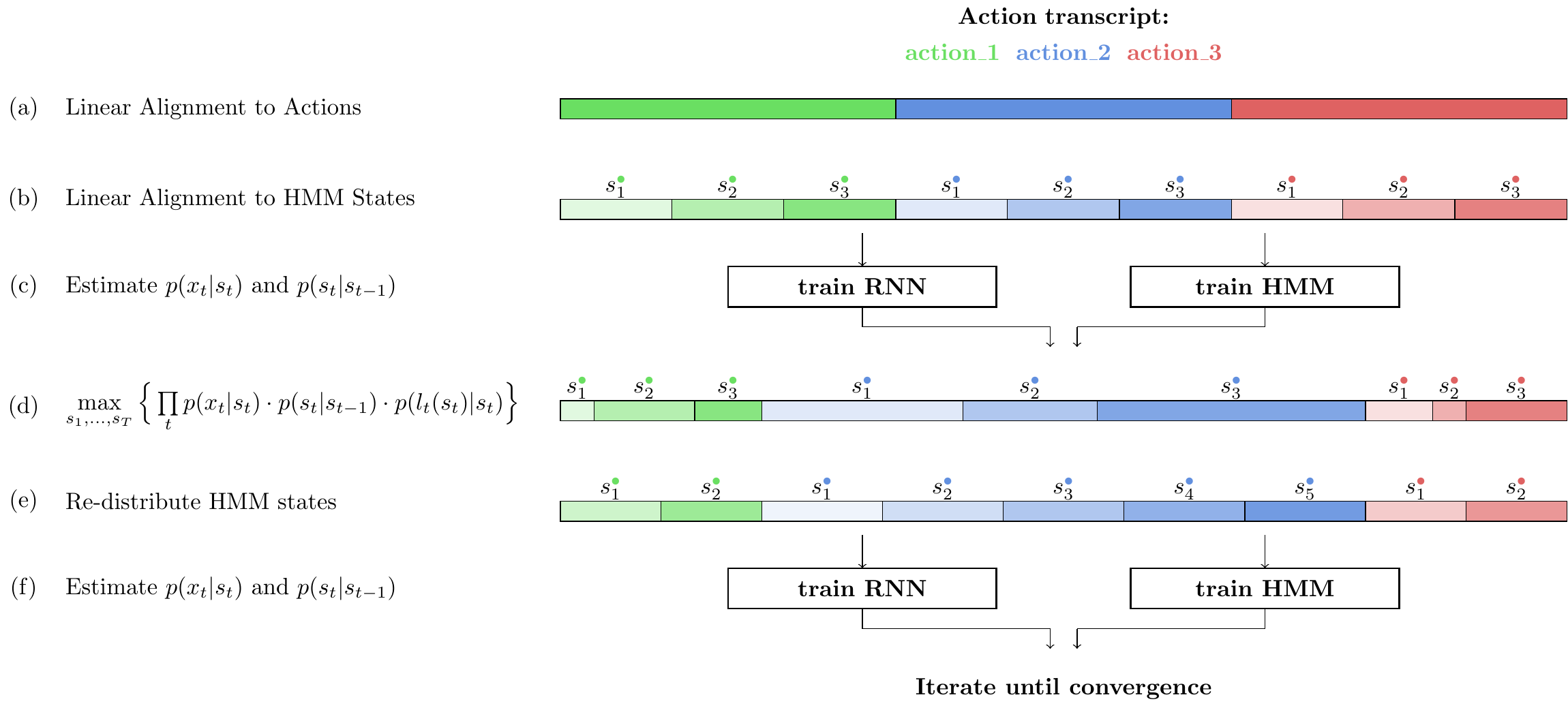}
    \caption{Training process of our model.
             Initially, each action is modeled with the same number of subactions
             and the video is linearly aligned to these subactions. Based on
             this alignment, the RNN is trained and used in combination
             with the HMMs to realign the video frames to the subactions.
             Eventually, the number of subactions per action is reestimated and the process
             is iterated until convergence.}
    \label{fig:train}
    \end{center}

\end{figure*}

\subsection{Training}
\label{sec:training}

The training of the model is done iteratively, altering between the recurrent neural network and the hidden Markov model training, and the alignment of frames to subactions via the hidden Markov model. The whole process is illustrated in Figure~\ref{fig:train}. We start with a linear segmentation and alignment of all training videos, train the respective RNN and HMM models and run an inference with the trained models which results in new frame boundaries for each action. We then redistribute the HMM states according to the new segmentation and repeat the training procedure several times until convergence is reached. In the following, we describe the two relevant steps, the initialization as well as the iterative training procedure in detail.  

\textbf{Initialization }
Each video is divided into $ N $ segments of equal size, where $ N $
is the number of action instances in the transcript (Figure~\ref{fig:train}a). 
Each action segment is further subdivided equally across the subactions (Figure~\ref{fig:train}b).
Note that this defines the mapping $ s(t) $ from frames to subactions. Additionally, each subaction should cover $ m $ frames of an action on average. We fix $ m $ to $10$ frames per subaction.
Thus, the initial number of subactions for each action is
\begin{align}
    \frac{\text{number of frames}}{\text{number of action instances} \cdot m}. \label{subactions}
\end{align}
Hence, initially each action is modeled with the same number of subactions. This can change during the following iterative optimization.

\textbf{Training }
The fine-grained RNN and the HMM are trained with the current mapping $ s_t $ as ground truth  (Figure~\ref{fig:train}c). Then, the RNN and HMM are applied to the training videos and a new alignment of frames to subactions (Figure~\ref{fig:train}d) is inferred given the new fine-grained probabilities $ p(x_t|s) $ from the RNN. The new alignment is obtained by finding the subaction mapping $ s_t $ that best explains the data:
\begin{align}
    \hat s_t &= \argmax_{s_t} \Big\{ p(\mathbf{x}_1^T|\mathbf{a}_1^N) \Big\} \nonumber \\
              &= \argmax_{s_t: \mathcal{A}(s_t) = \mathbf{a}_1^N} \Big\{ \prod_{t=1}^T p\big(x_t|s_t\big) \cdot p\big(s_t|s_{t-1}\big)  \cdot p \big(l_t(s_t)|s_t\big) \Big\}.
              \label{realign}
\end{align}
Again, Equation~\eqref{realign} can be efficiently computed using a Viterbi algorithm and we include our proposed length regularization in the recursive equation, \cf Equation~\eqref{viterbiRegularized}. For the case of training with sparse frame-level annotations, the resulting alignment is further refined as described in Section \ref{sec:alignmentWithFrameLabels}.

\textbf{Reestimation.} 
Once the realignment is computed for all training videos, the new average length of each action is computed and the number of subactions is re-estimated based on the updated average action lengths, which is computed as Equation~\eqref{subactions}, but for the entire action $ a $ instead of the state $ s $.
Correspondingly, there are now $ \mathrm{len}(a) / m $ subactions for action $ a $, which are
again uniformly distributed among the frames assigned to the action (Figure~\ref{fig:train}e). This new alignment is then used as current mapping and, as described in the beginning, the RNN can be trained with this new mapping and the HMM parameters can be updated (Figure~\ref{fig:train}f). As the mapping for each iteration requires a different number of outputs for the RNN, we train a new RNN-model for each iteration from scratch, initialized with random weights. These steps are iterated until convergence. 

\subsection{Stop criterion}
\label{sec:stopcriterion}

As the system iteratively approximates the optimal action segmentation on the training data, we define a stop criterion based on the overall amount of frame labels changed from one iteration to the succeeding one. Overall we stop if less than $5\%$ of the frames are assigned a new label or we reach a maximum of $15$ iterations.

\section{Setup}
\label{sec:setup}

\subsection{Datasets}
We evaluate the proposed approach on two different datasets.
The Breakfast dataset is a large scale dataset for hierarchical activity recognition and detection and comprises roughly about $ 4 $ million frames in $1,712$ clips and has an overall duration of $66.7$ hours. The dataset comprises $ 10 $ breakfast related tasks such as making tea but also complex activities such as the preparation of fried egg or pancake recorded with $ 52 $ different test persons in various kitchen environments. It features $ 48 $ action classes with a mean of $4.9$ instances per video. We follow the evaluation protocol as proposed by the authors in \cite{kuehne14language}.

The Hollywood extended \cite{bojanowski14weakly} dataset is an extension of the well known Hollywood dataset, featuring $937$ clips from different Hollywood movies with overall $787,720$ frames. The clips are annotated with two or more action labels resulting in $ 16 $ different action classes overall and a mean of $2.5$ action instances per clip. The authors propose a ten fold evaluation by selecting random clips. To allow for a better reproducibility of the results, we choose the last digit of the video number to define the splitting in the following evaluation.

Both datasets show a high heterogeneity, the first because of different persons, locations and camera viewpoints, the second because of the naturally high appearance variation of the movie sources. The two datasets differ in the mean and variance of the video length as well as in the  duration and variance of the annotated actions. This difference is important as it can be expected that modifications of the temporal modeling have a higher impact on data with higher temporal heterogeneity than on those with homogeneous temporal properties.

\subsection{Features}
For both datasets we computed the features as described in \cite{kuehne16end} using improved dense trajectories (IDT) and Fisher vectors (FVs). For the FV representation, we first reduce the dimensionality of the IDT features from 426 to 64 by PCA and sample $150,000$ randomly selected features to build a GMM with 64 Gaussians. The Fisher vector representation \cite{sanchez13image} for each frame is computed over a sliding window of 20 frames. Following \cite{perronnin10improving}, we apply power- and $\ell_2$-normalization to the resulting FV representation. Additionally, we reduce the final FV representation from $8,192$ to 64 dimensions via PCA to keep the overall video representation manageable and easier to process. 

\subsection{Alignment vs. Segmentation}
In case of weak learning of human actions two possible tasks can be considered to assess the accuracy of the proposed system, \emph{alignment} and \emph{segmentation}. In case of \emph{temporal action alignment} as \eg used by \cite{bojanowski14weakly, huang2016connectionist, kuehne2016weakly}, the test video and its respective transcript annotation is given and the task is to infer the segment boundaries based on the given order information. The accuracy in this case can be reported by mean over frames (MoF) as proposed by \cite{kuehne2016weakly} or by the Jaccard index (Jacc.), as in \cite{bojanowski14weakly}. In this case, the Jaccard index is computed as intersection over detection (IoD) and defined by $|G \cap D|/|D|$ with $G$ referring to the ground truth frames and $D$ referring to the detected action frames. 

In case of \emph{temporal action segmentation}, only the test video is given and the task is to infer the occurring actions as well as their respective frame boundaries.  We refer to temporal action segmentation as the combined video segmentation and classification. Thus, given a video without any further information, the task is to classify all frames according to their related action. This includes to infer which actions occur in the video, in which order they occur, and their respective start and end frames. This task is evaluated by the mean over frames (MoF) \cite{huang2016connectionist, kuehne2016weakly}.


\section{Evaluation}
\label{sec:evaluation}

We first evaluate the performance of the different components of our system, namely the GRU based classification, the subaction modeling, the length prior, and the semi-supervised setup for temporal action segmentation. We evaluate all tasks on the test set of the Breakfast dataset and report results as mean accuracy over frames (MoF). We iterate the system until the stop criterion as described in Section \ref{sec:stopcriterion} is reached.

\begin{figure*}[t]
    \centering
    \includegraphics[width=0.48\textwidth]{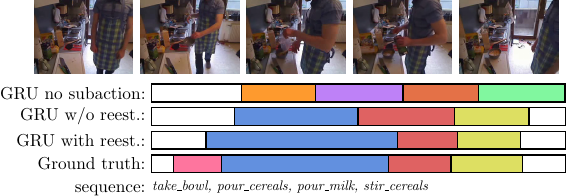}
    \hfill
    \vspace{3mm}
    \includegraphics[width=0.48\textwidth]{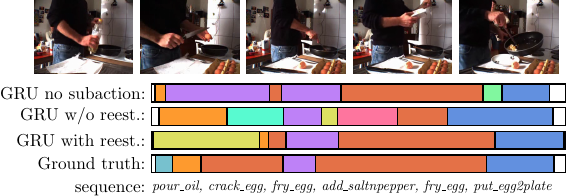}
    \caption{Example of temporal action segmentation for two samples from the Breakfast dataset showing the segmentation result for ´``preparing cereals'' and ``preparing friedegg''. Although the actions are not always correctly detected, there is still a reasonable alignment of detected actions and ground truth boundaries.}
    \label{fig:exp_seg}
\end{figure*}

\subsection{Evaluation of GRU-based model}
\label{sec:eval_GRU_rest}

\begin{table}[t] \footnotesize
    \centering
   \begin{tabularx}{0.45\textwidth}{Xrrrrr}
        \toprule
        Breakfast & Iter 1 & Iter 2 & Iter 3  & Iter 4  &  Iter 5   \\
        \cmidrule(lr){1-6}           
        \textit{GMM w/o reest.}   &  $  15.3 $ & $  23.3 $ & $  26.3 $ & $  27.0 $ & $  26.5 $  \\
        \textit{MLP w/o reest.}  &  $  22.4 $ & $  24.0 $ & $  23.7 $ & $  23.1 $ & $  20.3 $  \\
        \textit{GRU w/o reest.}  &  $  25.5 $ & $  29.1 $ & $  28.6 $ & $  29.3 $ & $  28.8 $  \\
        \bottomrule
    \end{tabularx}
    \vspace{3mm}
    \caption{Results for temporal action segmentation with GRU-based model compared to MLP-based model and GMM over five iterations. It shows that the MLP  and GMM are outperformed by the GRU-based model. Additionally the MLP-based model quickly starts to overfit whereas the GRU oscillates at a constant higher level.} 
    \label{tab:low_level_eval}
\end{table}

First, we evaluate the influence of the proposed fine grained RNN modeling.
In order to analyze the capability of capturing temporal context with the recurrent network, we compare it to a system where a multilayer perceptron (MLP) is used instead. The MLP only operates on frame level and does not capture temporal context as there is no recurrent connection involved. In order to provide a fair comparison to the recurrent model, we setup the MLP with a single hidden layer of rectified units  such that it has the same number of parameters as the recurrent network. We also look at the performance of standard GMM models, as they would be usually used in the context of HMMs. In this case, we follow the setup as described by \cite{kuehne16end}, using a single Gaussian distribution for each state of the model.

For this evaluation, we use a simplified version of the system without subaction reestimation or length prior to achieve comparable results after each iteration. We show results for the first five iterations in Table \ref{tab:low_level_eval}.
It becomes clear that GRUs outperform MLPs and GMMs, starting with $25.5\%$ for the initial recognition, and reaching up to $29.3\%$ after the fourth iteration. The MLP baseline stays continuously below this performance. Thus, it can be assumed that the additional information gained by recurrent connections in this context supports classification.
One can further see that the MLP reaches its best performance after the second iteration and then continuously decreases, whereas the GRU begins to oscillate around $29\%$, hinting that the MLP also starts to overfit at an earlier stage compared to the GRU. The GMMs are also performing better than the MLP but, with a maximum of $27.0\%$, do not reach the performance of the GRU.

\subsection{Analysis of the subaction modeling}
\label{sec:analysis_coarse}

 \begin{table}[t] 
    \centering
   \begin{tabularx}{0.45\textwidth}{Xr}
        \toprule
        Breakfast                            & Accuracy (Mof)  \\
        \midrule           
        \textit{GRU no subactions}        &  $  22.4 $  \\
        \textit{GRU w/o reestimation}    &  $  28.8 $  \\
        \textit{GRU + reestimation}      &  $  33.3 $  \\
        \midrule
       \textit{GRU + GT length}           &  $  51.3 $  \\
        \bottomrule
    \end{tabularx}
    \vspace{3mm}
    \caption{Results for temporal action segmentation on the Breakfast dataset comparing accuracy of the proposed system (GRU + reestimation) to the accuracy of the same architecture without subactions (GRU no subactions) and to the architecture with subclasses but without reestimation. } 
    \label{tab:reest_eval}
\end{table}

\begin{figure}[t]
    \centering
    \includegraphics[width=0.45\textwidth]{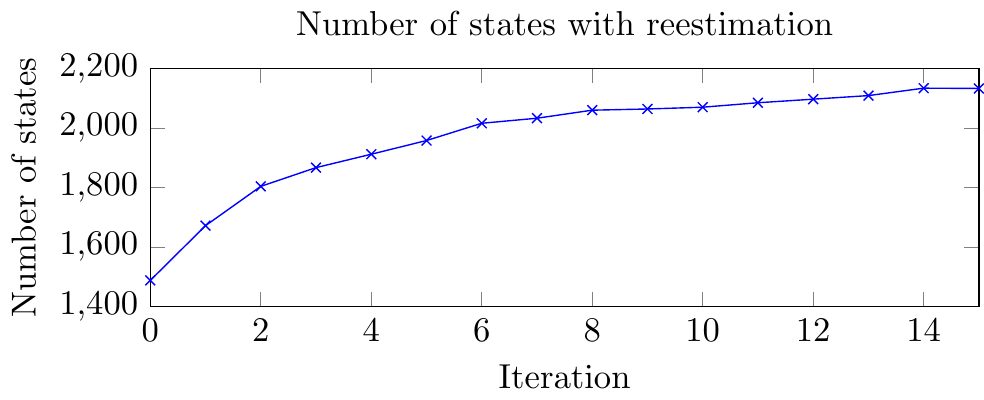}
    \caption{Evolution of number of states for the model with state reestimation. The number of states increases in the first five iterations and converges after ten iterations. }
    \label{fig:number_of_states}
\end{figure}

Second, we regard the properties of the proposed subaction modeling. We therefore compare the proposed system with the results of the same setting, but without further subdividing actions into subactions (GRU no subactions, Table~\ref{tab:reest_eval}). Additionally, we regard results of the system without reestimation of subactions during optimization (GRU w/o reestimation, Table~\ref{tab:reest_eval}). For the system without reestimation, we follow the initial steps as shown in Figure \ref{fig:train}, thus, we linearly segment the videos according to the number of actions, generate an initial subaction alignment, train the respective subaction classes, and realign the sequence based on the RNN output. But, opposed to the setup with reestimation, we omit the step of reestimating the number of subclasses and the following alignment. Instead, we just use the output of the realignment (see Figure~\ref{fig:train}d) to retrain the classifier and iterate the process of training, alignment, and re-training. Thus, the number of subclasses is constant and the coarse model is not adapted to the overall estimated length of the action class. 

Finally, we compare to an approach in which we use the ground truth boundaries to compute the mean length of an action class and set the number of subactions based on the mean ground truth length (GRU + GT length, Table~\ref{tab:reest_eval}).  Here, all action classes are still uniformly initialized, but longer action classes are divided into more subactions than shorter ones. We include this scenario as it models the performance in case that the optimal number of subaction classes would be found. We again use a simplified version of the system without length prior to achieve comparable results.

Table \ref{tab:reest_eval} shows that the performance without subactions is 
significantly below all other configurations, supporting the idea that subaction 
modeling in general helps recognition in this scenario. 
The model with subactions, but without reestimation, improves over the single 
class model, but is still below the system with subaction reestimation. Compared 
to that, the model with subaction reestimation performs 5\% better. 
We ascribe the performance increase of the reestimated model to the fact that a 
good performance is highly related to the correct number of subactions, thus to 
a good length representation of the single actions.  The impact of the number of 
subactions becomes clear, when considering the results when the ground truth 
action lengths are used. The performance of the same system, just with different 
numbers of subactions, increases by almost $20\%$. 
This effect becomes also visible in the qualitative results as shown in Figure 
\ref{fig:exp_seg}.  Comparing the results of the three configurations - without 
subactions, without restimation, and with reestimation - to the ground truth 
annotations, it shows that, although the overall sequence is not always 
correctly inferred by the models, the system with reestimation finds a good 
alignment compared to the ground truth frame boundaries.
We also regard how the overall number of subactions changes by reestimating 
after each iteration as shown in Figure \ref{fig:number_of_states}. It becomes 
visible that the number of subactions mainly increases in the beginning and 
starts to converge after five to ten iterations.
approximated.
ground truth distribution by XX\% percent, w as the the state distribution of 
the reestimated model e.g. after iteration eight deviates only by XX\%.

\begin{figure*}[t]
    \centering
    \includegraphics[width=0.23\textwidth]{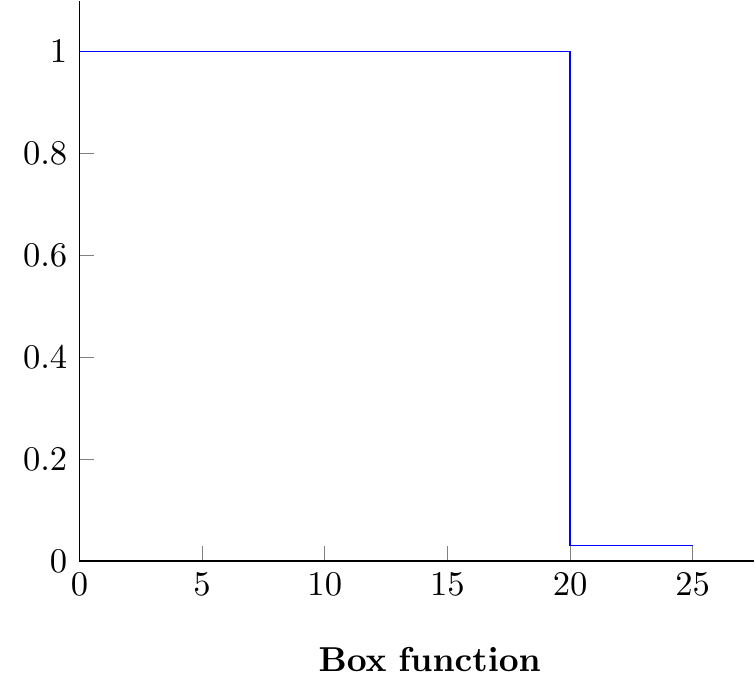}
    \vspace{3mm}
    \includegraphics[width=0.23\textwidth]{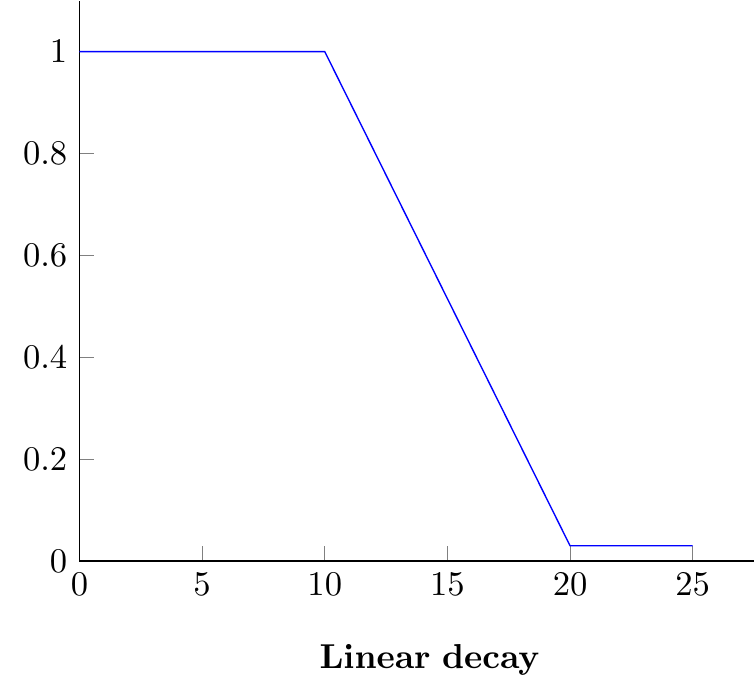}
    \vspace{3mm}
    \includegraphics[width=0.23\textwidth]{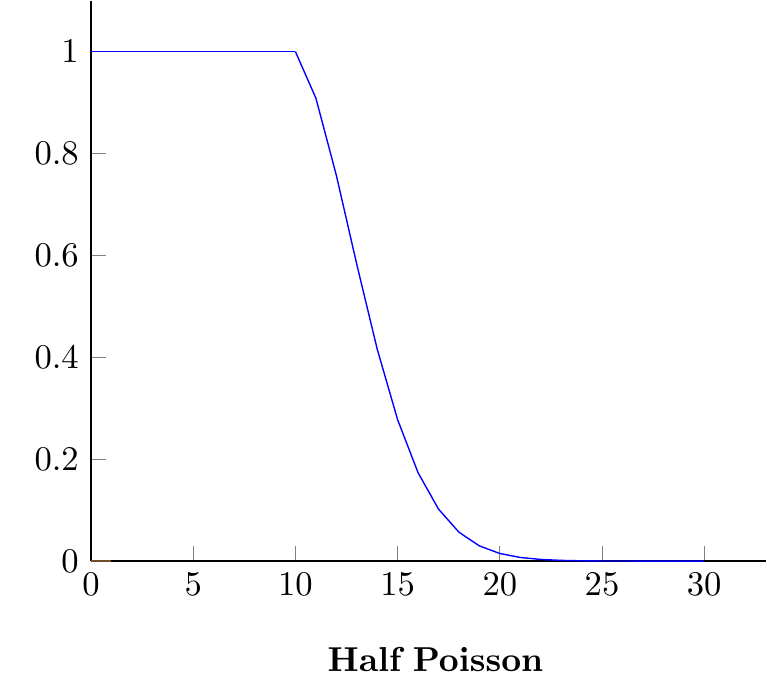}
    \vspace{3mm}
    \includegraphics[width=0.23\textwidth]{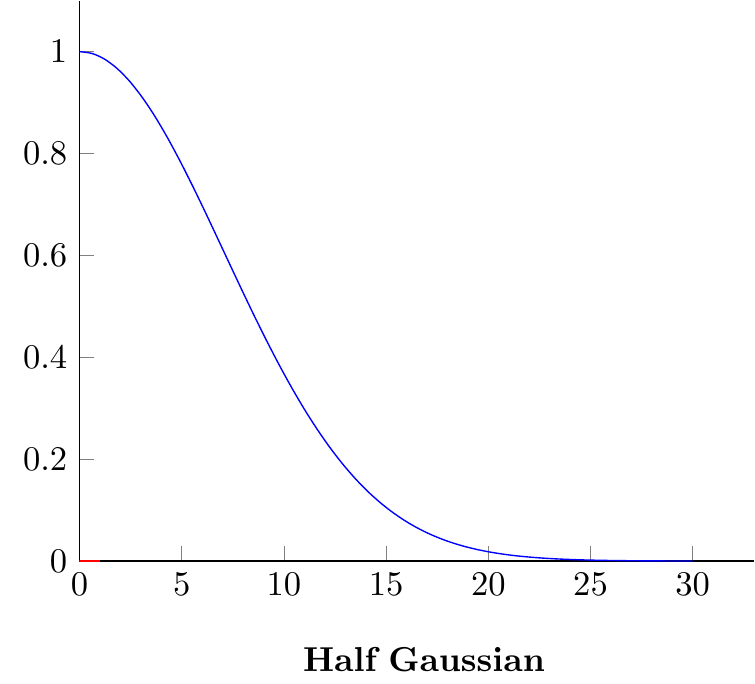}
    \caption{Overview of evaluated length models showing a simple box function, 
a linear decay function, a half Poisson decay and a half Gaussian function for a 
subaction with a mean length of $ 10 $ frames. See Appendix for formulas of the 
functions.}
    \label{fig:length_model}
\end{figure*}

\subsection{Analysis of length prior}
\label{sec:eval_lengthprior}

\begin{figure}[t]
    \centering
    \includegraphics[width=0.43\textwidth]{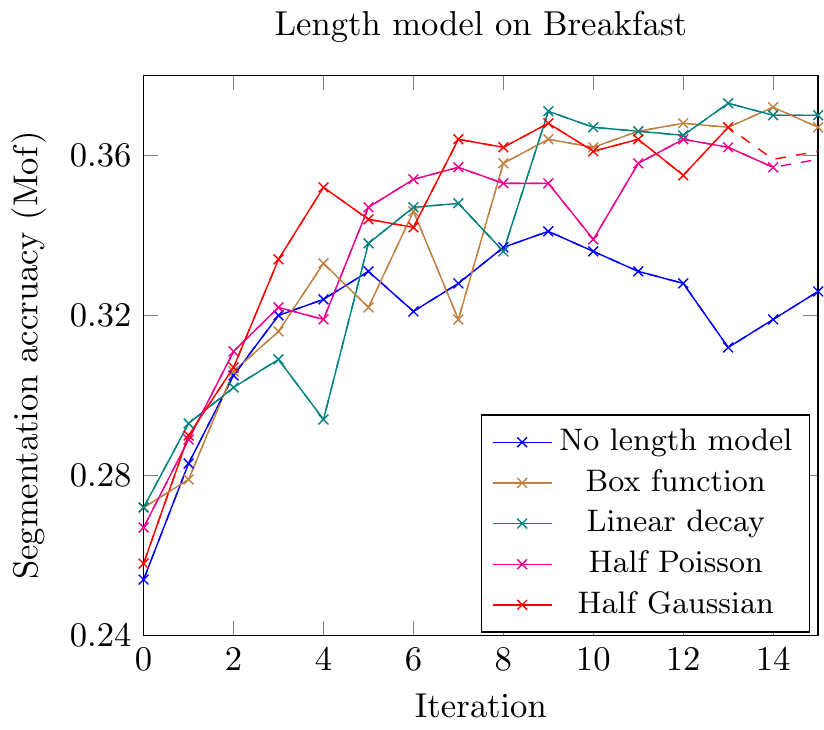}
    \caption{Results for temporal segmentation with different length models on the Breakfast dataset over 15 iterations. Solid lines show the results until the proposed stop criterion is reached. Dashed lines show the results after the stop criterion.}
    \label{fig:Breakfast_LM}
\end{figure}


 \begin{table}[t] 
    \centering
   \begin{tabularx}{0.45\textwidth}{Xrr}
        \toprule
        Length model                & Breakfast(MoF)  & Hollywood Ext.(IoU)  \\
        \midrule           
        \textit{No length model}        &  $  32.6 $ & $ 11.5 $ \\
        \midrule           
        \textit{Box function}        &  $  36.7 $ & $ 9.9 $ \\
        \textit{Linear decay}        &  $  37.0 $  & $ 10.5 $\\
        \textit{Half Poisson}        &  $  35.7 $  & $ 11.1 $\\
       \textit{Half Gaussian}           &  $  36.7 $ & $ 12.3 $ \\
        \bottomrule
    \end{tabularx}
    \vspace{3mm}
    \caption{Results of temporal action segmentation for different length prior functions on the Breakfast and the Hollywood Extended dataset. All results are based on a stop criterion of $5\%$ frame change rate during alignment or a maximum of 15 iterations. } 
    \label{tab:lm_eval}
\end{table}

In a next step, we analyze the impact of the length prior on the overall system. In terms of decay functions, we evaluate four different types of functions as shown in Figure \ref{fig:length_model}, a simple box function, a linear decay function, a half Poisson decay and a half Gaussian function.

First, we evaluate the overall performance of the different functions on the system. For all measures, we use the stop criterion as described in Section \ref{sec:stopcriterion}. As Table \ref{tab:lm_eval} shows, the length model mainly improves the results for the Breakfast dataset. All length models are doing better than the original system without length model in this case, with a best accuracy of $37.0\%$ reached by the linear decay function. This is further supported by the evaluation of the performance during training as the plot in Figure~\ref{fig:Breakfast_LM} shows. Here, the solid line shows the segmentation accuracy of the models after each iteration until the stop criterion is reached. After that, additional results are displayed by a dashed line. It shows that all four functions significantly outperform the system without length prior with best results at $36.7\%$ for box and $35.7\%$ and  $36.7\%$ for half Poisson and Gaussian. 

The impact of the length models on the Hollywood Extended dataset is smaller than on the Breakfast dataset as shown in Table \ref{tab:lm_eval}. This behaviour can be based on the fact that the actions in the Hollywood Extended dataset are usually shorter and the action classes have a lower temporal variance compared to the Breakfast dataset. 
On Hollywood extended, all action classes usually have a consistent mean frame length, where as in case of Breakfast, the mean length of action classes significantly varies. Thus, the benefit of using length models increases with the heterogeneity of the target action classes. Therefore it can be expected that a length model has less impact in this case than for datasets with high temporal variance among the action classes.

Overall, based on the numbers in Table \ref{tab:lm_eval}, it can be stated that the half Gaussian function gives the most consistent improvement for both datasets. We therefore use a length prior with a half Gaussian function for the following experiments.


\subsection{Semi supervised learning including sparse frame-level annotation}

Finally, we evaluate the behaviour of the proposed system in a semi-supervised setup including sparse frame-based annotation. We follow the idea presented in \cite{huang2016connectionist} of using a fraction of all frames as additional information during training. We again report all results for the segmentation task on the Breakfast dataset. We start with a fraction of $0.25\%$ of all annotated frames which roughly corresponds to one frame per action instance in the dataset. We uniformly select the frame annotation from the ground truth dataset. Note that not all action instances will have a respectively labeled frame and we do not get any information about the segment boundaries by this type of labeling. 

\begin{figure}[t]
    \centering
    \includegraphics[width=0.43\textwidth]{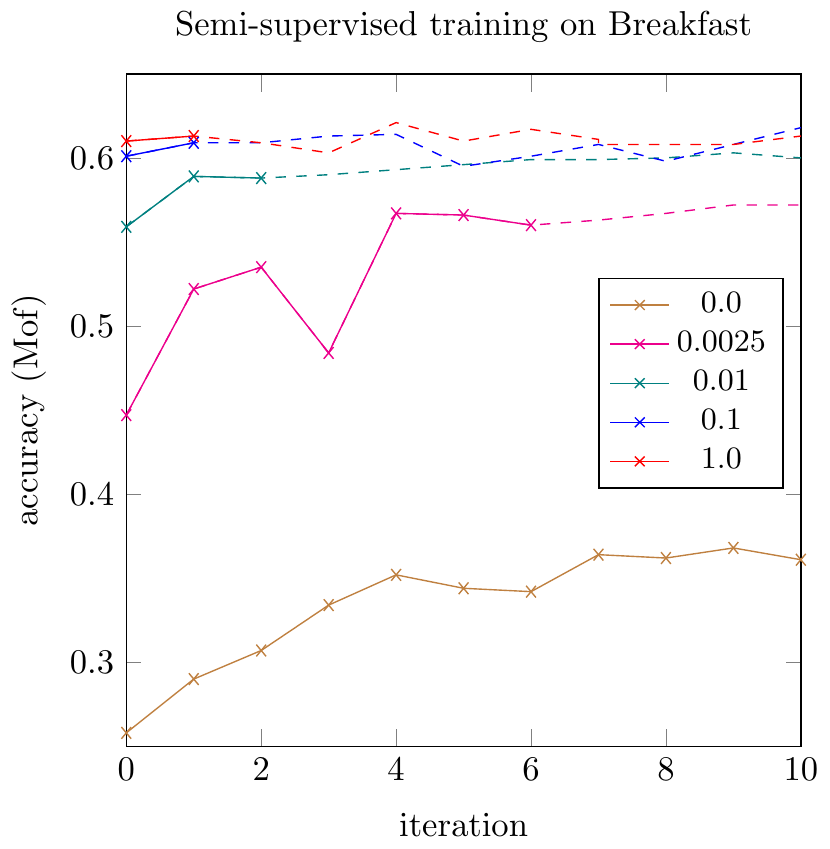}
    \caption{Results of temporal action segmentation with semi supervised training on the Breakfast dataset for 10 iterations. Solid lines show the results until the proposed stop criterion is reached, dashed lines show the results after the stop criterion. It shows that even small fractions of annotated frames can significantly improve the overall performance of the system. }
    \label{fig:Breakfast_SemiSupervised}
\end{figure}

Results for the sparse frame-level annotation are shown in Table \ref{tab:partial_annotation}.  It shows that even a small fraction of annotated frames ($0.25\%$) helps to improve the overall accuracy on the Breakfast dataset by almost $20\%$ reaching $56.0\%$. It further shows that with only $1\%$ of annotated frames, the result is already getting close to the fully supervised setting of $61.0\%$. 
Additionally, we observe a faster convergence behaviour when including more frame-level annotation. The system without frame-level annotation usually needs 12-15 iterations to meet the respective convergence criterion. As can be seen in Figure \ref{fig:Breakfast_SemiSupervised} in case of frame level supervision, we reach convergence already after two to four iteration steps, as the overall frame alignment obviously shifts less the more frame information is available. This supports the idea that a partial annotation of frames might be a valid alternative to the time consuming, fully supervised labeling.

The same behaviour at a smaller scale is also visible for the Hollywood Extended dataset (see Table \ref{tab:partial_annotation}). Here, we mainly see an increase when $1\%$ or more frames are annotated. It can be assumed that the smaller improvement is based on two points. First, the weakly supervised accuracy of $12.3\%$ of the Hollywood Extended dataset is already close to the fully supervised setting with $13.7\%$.  Thus, the increase can only be within this margin. Second, we again observe that the additional information during training mainly helps for a better temporal alignment of the data. Thus, it can be expected that the influence on datasets with higher temporal variance is stronger than for those with lower temporal variance.

 \begin{table}[t] 
    \centering
   \begin{tabularx}{0.45\textwidth}{Xrr}
        \toprule
        Fraction                            & Breakfast(Mof)  & Hollywood Ext.(IoU)  \\
        \midrule           
        \textit{0.0}          &  $  36.7 $ & $ 12.3 $ \\
        \midrule           
        \textit{0.0025}       &  $  56.0 $ & $ 12.3 $ \\
        \textit{0.01}         &  $  58.8 $ & $ 13.1 $\\
        \textit{0.1}          &  $  60.9 $ & $ 13.3 $\\
        \textit{1}            &  $  61.3 $ & $ 13.7 $ \\
        \bottomrule
    \end{tabularx}
    \vspace{3mm}
    \caption{Results for temporal action segmentation wtih semi supervised learning on the Breakfast and the Hollywood Extended dataset with a half Gaussian length prior. The fraction indicates how many frames of the data were labeled. A fraction of 1 corresponds to a fully supervised setup.  All results are based on the stop criterion of $5\%$ frame change rate during alignment or a maximum of 15 iterations.  } 
    \label{tab:partial_annotation}
\end{table}


\subsection{Comparison to State-of-the-Art}

 \begin{table}[t] \footnotesize
    \centering
   \begin{tabularx}{0.45\textwidth}{Xr}
        \toprule
        \multicolumn{2}{c}{\textbf{Breakfast}}\\
        \midrule
         Model &  Accuracy (Mof)  \\
        \midrule           
        OCDC \cite{bojanowski14weakly}*   &  $ 8.9 $  \\
        HTK \cite{kuehne2016weakly}  &  $ 25.9 $  \\
        ECTC \cite{huang2016connectionist}  &  $ 27.7 $  \\
        TCFPN \cite{Ding2018weakly} &  $ \mathbf{38.4} $  \\ 
        \midrule 
        GRU-RNN \cite{richard2017weakly}  &  $ 33.3 $  \\
        GRU + length prior &  $ 36.7 $  \\
       \bottomrule
       \\
        
       \\
        \toprule
        \multicolumn{2}{c}{\textbf{Hollywood Extended}}\\
        \midrule
         Model &  Jacc (IoU)  \\
        \midrule           
        HTK \cite{kuehne2016weakly}  &  $ 8.6 $  \\
        TCFPN \cite{Ding2018weakly} &  $ \mathbf{12.6} $  \\ 
        \midrule 
        GRU-RNN \cite{richard2017weakly}  &  $ 11.9 $  \\
        GRU  + length prior  &  $ 12.3 $  \\
       \bottomrule
    \end{tabularx}
    \vspace{3mm}
    \caption{Comparison of temporal action segmentation performance for GRU based weak learning with other approaches. For the Breakfast dataset, we report performance as mean over frames (Mof), for Hollywood extended, we measure the Jaccard index as intersection over union for this task (*from \cite{huang2016connectionist}).} 
    \label{tab:sota1}
\end{table}

\textbf{Temporal Action Segmentation}
We compare our system to four different approaches published for this task: The first is the Ordered Constrained Discriminative Clustering (OCDC) proposed by Bojanovski \etal~\cite{bojanowski14weakly}, which has been introduced on the Hollywood extended dataset. Second, we compare against the HTK system used by Kuehne \etal~\cite{kuehne2016weakly}, third against the Extended Connectionist Temporal Classification (ECTC) by Huang \etal~\cite{huang2016connectionist} and fourth against the temporal
convolutional feature pyramid network (TCFPN) by Ding and Xu \cite{Ding2018weakly}. We further compare against a previous version of this system without lenght model \cite{richard2017weakly}.

For the Breakfast dataset, we follow the evaluation protocol of \cite{kuehne14language} and \cite{huang2016connectionist} and report results as mean accuracy over frames over four splits. For the Hollywood Extended dataset, we follow the evaluation protocol of \cite{kuehne2016weakly} and report the Jaccard index (Jacc.) as intersection over union (IoU) over 10 splits. 

Results are shown in Table~\ref{tab:sota1}. One can see that both GRU systems show a good performance, and that the proposed system outperforms most current approaches on the evaluated datasets. Only the recently released TCFPN reaches better results on this task.

 \begin{table}[t] \footnotesize
  \centering
   \begin{tabularx}{0.45\textwidth}{Xr}
        \toprule
        \multicolumn{2}{c}{\textbf{Breakfast}}\\
        \midrule
         Model &  Jacc. (IoD)  \\
        \midrule           
        OCDC \cite{bojanowski14weakly}  &  $ 23.4 $  \\
        HTK \cite{kuehne2016weakly}  &  $ 42.4 $  \\
        TCFPN \cite{Ding2018weakly} &  $ 52.3 $  \\ 
        \midrule 
        GRU-RNN \cite{richard2017weakly}  &  $ 47.3 $  \\
        GRU  + length prior &  $ \mathbf{52.4} $  \\
       \bottomrule 
       \\
        
       \\
        \toprule
        \multicolumn{2}{c}{\textbf{Hollywood Extended}}\\
        \midrule
         Model &  Jacc. (IoD)  \\
        \midrule           
        OCDC  \cite{bojanowski14weakly}** &  $ 43.9 $  \\
        HTK \cite{kuehne2016weakly}**  &  $ 42.4 $  \\
        ECTC \cite{huang2016connectionist}** &  $ 41.0 $  \\ 
        TCFPN \cite{Ding2018weakly} &  $ 39.6 $  \\ 
        \midrule 
        GRU-RNN \cite{richard2017weakly}  &  $ \mathbf{46.3} $  \\
        GRU  + length prior &  $ 46.0 $  \\
       \bottomrule
    \end{tabularx}
    \vspace{3mm}
    \caption{Results for temporal action alignment on the test set of the Breakfast and the Hollywood extended dataset reported as Jaccard index of intersection over detection (IoD)(**results obtained from the authors).} 
    \label{tab:sota2}
\end{table}

\noindent\textbf{Temporal Action Alignment }
We also address the task of action alignment. We assume that given a video and a sequence of temporally ordered actions, the task is to infer the respective boundaries for the given action order. We report results for the test set of Breakfast as well as for the Hollywood Extended dataset based on the Jaccard index (Jacc.) computed as intersection over detection (IoD) as proposed by \cite{bojanowski14weakly}. The results are shown in Table~\ref{tab:sota2}.

The proposed approach outperforms current state-of-the-art approaches. It also shows that the system without length model performs slightly better than the system with length model for the alignment in case of Hollywood Extended. As already discussed in Section~\ref{sec:eval_lengthprior}, the differences between the proposed approach with and without length model are marginal on this dataset. 

\noindent\textbf{Fully supervised classification }
We finally evaluate the approach in a fully supervised setting and compare it to other proposed approaches. Here, we compute the mean length from the training annotations directly and use it to determine the number of states as well as the length parameter of the prior function. 

 \begin{table}[t] \footnotesize
  \centering
   \begin{tabularx}{0.45\textwidth}{Xr}
        \toprule
        \multicolumn{2}{c}{\textbf{Breakfast}}\\
        \midrule
         Model &  MoF  \\
        \midrule           
        HMM-BOW \cite{kuehne14language}  &  $ 28.8 $  \\
        HMM-FV \cite{kuehne16end}        &  $ 56.3 $  \\
        TCFPN \cite{Ding2018weakly}        &  $ 52.0  $  \\
        \midrule 
        GRU  w/o length prior      &  $ \mathbf{60.2} $  \\
        GRU  + length prior      &  $ \mathbf{61.3} $  \\
       \bottomrule
    \end{tabularx}
    \vspace{3mm}
    \caption{Results for fully supervised temporal action segmentation on the Breakfast dataset (MoF).} 
    \label{tab:sota_fully_sup}
\end{table}

Table \ref{tab:sota_fully_sup} shows that the system clearly outperforms previous approaches. Since the apporaches \cite{kuehne14language} and \cite{kuehne16end} have a similar hierarchical structure as the presented system and similar features \cite{kuehne16end}, we can assume that the main improvement can be attributed to the underlying GRU models. This is consistent with the findings in Section \ref{sec:eval_GRU_rest}, where it shows that the proposed GRUs improve fine-grained frame-based classification compared to other approaches. Additionally, it shows that the length model also improves the segmentation accuracy in case of fully supervised training. This is important as in this case, we can assume that all other temporal factors, such as the number of states are already optimal. Thus, even in this case, a temporal prior can improve the overall recognition of the system.


\section{Conclusion}
\label{sec:conclusion}

We presented an approach for weakly, semi and fully supervised learning of human actions based on a combination of a discriminative representation of subactions modeled by a recurrent neural network and a coarse probabilistic model to allow for a temporal alignment and inference over long sequences. Although the system itself shows already good results, the performance is significantly improved by approximating the number of subactions for the different action classes and by adding a length prior formulation to the overall system. Accordingly, we combine the length model with the adaptation of the number of subaction classes by iterating realignment and reestimation during training. The resulting model shows a competitive performance on various weak learning tasks such as temporal action segmentation and action alignment on two standard benchmark datasets.

\appendices
\section{}

In the following we report the formulas used for the four different length models evaluated in our work as well as their basic properties. Note that we abbreviate $ l_t(s_t) $ by $ l_t $ for better readability. Also note that each distribution is normalized in a way that $ \max\{\tilde p (l|s)\} = 1 $, \ie for each of the monotonically decreasing functions, the highest value is one. Although the models are not a strict probability distribution anymore, the normalization simplifies the formulas and sums up to a constant in the Viterbi decoding, not affecting the overall outcome. We set the $\epsilon$ in all functions to $0.001$. All models are also displayed in Figure \ref{fig:length_model}.

\subsection*{Box Function}

\begin{align}
    \tilde p \big(l_t|s_t\big) = \begin{cases}
               1 , & l_t \leq 2 \cdot \mathrm{len}(s_t)\\
               \epsilon, & l_t > 2 \cdot \mathrm{len}(s_t)\\
            \end{cases}
\end{align}

The box function is considered as the basic representation of a length model. In this case the length prior does not influence the inference up to the point that twice the mean length of the respective state is reached. After that, the overall probability is multiplied with a given $\epsilon$ and thus marginalized, so that the respective state is not used anymore.

\subsection*{Linear Decay}

\begin{align}
    \tilde p \big(l_t|s_t\big) = \begin{cases}
               1 , & l_t \leq \mathrm{len}(s_t)\\
               1 - \frac{l_t - \mathrm{len}(s_t)}{\mathrm{len}(s_t)} , & l_t > \mathrm{len}(s_t) \land l_t < 2 \cdot \mathrm{len}(s_t)\\
               \epsilon, & l_t \geq 2 \cdot \mathrm{len}(s_t)\\
            \end{cases}
\end{align}

The linear decay function can be seen as an extension of the box function. Here, the length prior is fix up to the point that the mean length of the respective state is reached. Then, the overall length prior linearly decreases, punishing longer states more than shorter ones. After twice the mean length is reached, the length prior is set to $\epsilon$ and the overall probability is thus marginalized, so that the respective state is not used anymore.

\subsection*{Half Poisson}

\begin{align}
    \tilde p \big(l_t|s_t\big) = \begin{cases}
               1 , & l_t \leq \mathrm{len}(s_t)\\
               \mathrm{const} \cdot \frac{\mathrm{len}(s_t)^{l_t}}{l_t!} e^{-\mathrm{len}(s_t)}, & l_t > \mathrm{len}(s_t)\\
            \end{cases}
\end{align}
For the Half Poisson, $ \mathrm{const} $ is a normalization factor such that $ \max_l \big\{ \tilde p \big(l|s_t\big) \big\} = 1 $.
The half Poisson model in the here proposed case also starts with a plateau and is fix up to the point that the mean length of the respective state is reached. After that we consider the right-half of the Possion distribution of the respective state. We choose this combination as we want to model the discrete distribution of state lengths, and at the same time, ensure a monotonically decreasing function. 

\subsection*{Half Gaussian}

\begin{align}
    \tilde p \big(l_t|s_t\big) = e^{-\frac{(l_t-\mu)^2}{\sigma^2}}
\end{align}

Closely related to the half Poisson model is the half Gaussian model. Here, the property of a monotonically decreasing function is implicitly ensured by setting $\mu = 0$ and $\sigma = \mathrm{len}(s_t)$.

\ifCLASSOPTIONcompsoc
  \section*{Acknowledgments}
\else
  \section*{Acknowledgment}
\fi

The work has been financially supported by the DFG projects KU 3396/2-1 (Hierarchical Models for Action Recognition and Analysis in Video Data) and GA 1927/4-1 (DFG Research Unit FOR 2535 Anticipating Human Behavior) and the ERC Starting Grant ARCA (677650). This work has been supported by the AWS Cloud Credits for Research program.




\bibliographystyle{IEEEtran}
\bibliography{references}
%

%
\vspace{-7mm}

\begin{IEEEbiography}[{\includegraphics[width=1in,height=1.25in,clip,
keepaspectratio]{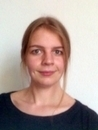}}]{Hilde Kuehne}
Dr. Hilde Kuehne obtained her Diploma in Computer Science from the University of Koblenz-Landau in 2006 and got her doctoral degree in engineering form the Karlsruhe Institute of Technology (KIT) in 2014. From 2013 to 2016 she worked as a senior researcher at the Fraunhofer Institute for Communication, Information Processing and Ergonomics FKIE. In 2016, she joined the Computer Vision Group headed by Prof. Gall at the Institute of Computer Science at the Universtity of Bonn. 
\end{IEEEbiography}
\vspace{-7mm}

\begin{IEEEbiography}[{\includegraphics[width=1in,height=1.25in,clip,
keepaspectratio]{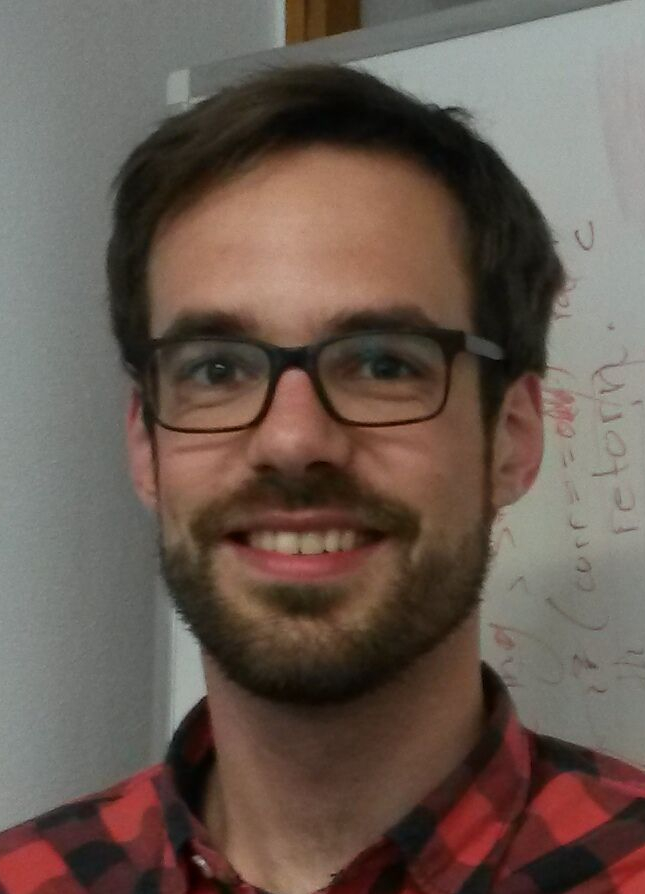}}]{Alexander Richard}
Alexander Richard received his Masters degree in Computer Science from RWTH Aachen University in 2014, specializing on automatic speech recognition. In 2014, he joined the Computer Vision Group of Prof. Gall at the Institute of Computer Science at the University of Bonn as a doctoral student researcher. His research focuses on video analysis and automatic detection and classification of human actions.
\end{IEEEbiography}
\vspace{-7mm}

\begin{IEEEbiography}[{\includegraphics[width=1in,height=1.25in,clip,
keepaspectratio]{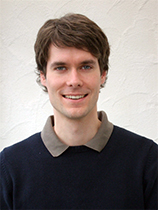}}]{Juergen Gall}
Juergen  Gall  obtained  his  B.Sc. and his Masters degree in mathematics from the University  of  Wales  Swansea  (2004)  and  from
the University of Mannheim (2005). In 2009, he obtained a Ph.D. in computer science from the Saarland University and the Max Planck Institut f\"ur Informatik. He was a postdoctoral researcher
at the Computer Vision Laboratory, ETH Zurich, from 2009 until 2012 and senior research scientist at the Max Planck Institute for Intelligent Systems  in  T\"ubingen  from  2012  until  2013.  Since 2013, he is professor at the University of Bonn and head of the Computer Vision Group.
\end{IEEEbiography}

\vfill




\end{document}